\documentclass[letterpaper]{article} 
\usepackage{aaai24}  
\usepackage{times}  
\usepackage{helvet}  
\usepackage{courier}  
\usepackage[hyphens]{url}  
\usepackage{graphicx} 
\usepackage{natbib}  
\usepackage{caption} 
\usepackage{algorithm}
\usepackage{algorithmic}
\usepackage{times}
\usepackage{helvet}
\usepackage{courier}
\usepackage{microtype}
\usepackage{amsfonts}
\usepackage{amssymb}
\usepackage{multirow}
\usepackage{graphicx}
\usepackage{amssymb}
\usepackage{natbib}
\usepackage{amsmath}
\usepackage{appendix}
\usepackage{subfigure}
\usepackage{booktabs}
\newtheorem{lemma}{Lemma}
\usepackage{algorithm}
\usepackage{algorithmic}

\usepackage{newfloat}
\usepackage{listings}
\DeclareCaptionStyle{ruled}{labelfont=normalfont,labelsep=colon,strut=off} 
\floatstyle{ruled}
\newfloat{listing}{tb}{lst}{}
\floatname{listing}{Listing}
\title{Transfer and Alignment Network for Generalized Category Discovery}
\author{
    Wenbin An\textsuperscript{\rm 1,3}, Feng Tian\textsuperscript{\rm 2,3}\footnotemark[1], Wenkai Shi\textsuperscript{\rm 1,3}, Yan Chen\textsuperscript{\rm 2}, Yaqiang Wu\textsuperscript{\rm 4}, Qianying Wang\textsuperscript{\rm 4}\thanks{Corresponding author.}, Ping Chen\textsuperscript{\rm 5}\\
}
\affiliations{
    \textsuperscript{\rm 1}School of Automation Science and Engineering, Xi'an Jiaotong University \\
    \textsuperscript{\rm 2}School of Computer Science and Technology, Xi'an Jiaotong University \\
    \textsuperscript{\rm 3}National Engineering Laboratory for Big Data Analytics \\
    \textsuperscript{\rm 4}Lenovo Research \\
    \textsuperscript{\rm 5}Department of Engineering, University of Massachusetts Boston\\
    \{wenbinan,kkkkkkai0611\}@stu.xjtu.edu.cn, \{fengtian,chenyan\}@mail.xjtu.edu.cn \\
    \{wuyqe,wangqya\}@lenovo.com,Ping.Chen@umb.edu
}

\usepackage{bibentry}

\begin{document}

\maketitle
\begin{abstract}
Generalized Category Discovery (GCD) is a crucial real-world task that aims to recognize both known and novel categories from an unlabeled dataset by leveraging another labeled dataset with only known categories. 
Despite the improved performance on known categories, current methods perform poorly on novel categories. 
We attribute the poor performance to two reasons: \textit{biased knowledge transfer} between labeled and unlabeled data and \textit{noisy representation learning} on the unlabeled data. The former leads to unreliable estimation of learning targets for novel categories and the latter hinders models from learning discriminative features.
To mitigate these two issues, we propose a \textit{\textbf{T}ransfer and \textbf{A}lignment \textbf{N}etwork} (TAN), which incorporates two \textit{knowledge transfer mechanisms} to calibrate the biased knowledge and two \textit{feature alignment mechanisms} to learn discriminative features.
Specifically, we model different categories with prototypes and transfer the prototypes in labeled data to correct model bias towards known categories.
On the one hand, we pull instances with known categories in unlabeled data closer to these prototypes to form more compact clusters and avoid boundary overlap between known and novel categories. On the other hand, we use these prototypes to calibrate noisy prototypes estimated from unlabeled data based on category similarities, which allows for more accurate estimation of prototypes for novel categories that can be used as reliable learning targets later.
After knowledge transfer, we further propose two feature alignment mechanisms to acquire both instance- and category-level knowledge from unlabeled data by aligning instance features with both augmented features and the calibrated prototypes, which can boost model performance on both known and novel categories with less noise.
Experiments on three benchmark datasets show that our model outperforms SOTA methods, especially on novel categories. Theoretical analysis is provided for an in-depth understanding of our model in general.
Our code and data are available at \url{https://github.com/Lackel/TAN}.

\end{abstract}

\section{Introduction}
Despite remarkable breakthroughs achieved by modern deep learning systems, the majority of models are designed under a close-world setting, based on the assumption that training and test data are from the same set of pre-defined categories \citep{openworld}. However, many practical problems such as intent detection \citep{thu2020} and relation extraction \citep{hogan2023open} are open-world, where the well-trained models may encounter unlabeled data containing unseen novel categories.
To meet the open-world demands, Generalized Category Discovery (GCD) was widely studied in both NLP \citep{thu2021,dpn} and CV fields \citep{gcd,simple}. 
GCD requires models to recognize both known and novel categories from encountered unlabelled data based on a set of labeled data with only known categories, which can adapt models to the increasing number of categories without any additional labeling cost.

\begin{figure}
\centering
\includegraphics[width=7.5cm, height=3.8cm]{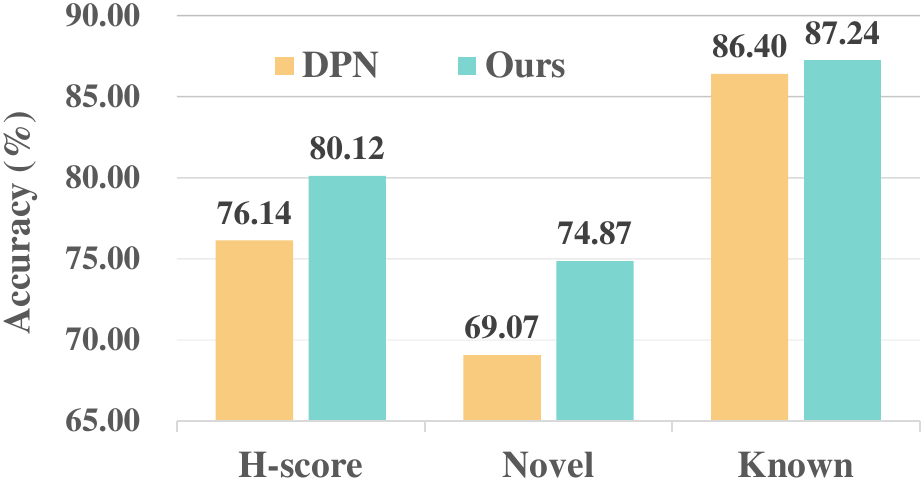}
\caption{Average accuracy on three benchmark datasets compared to the SOTA method DPN. Accuracy for novel categories (Novel), known categories (Known) and their harmonic mean (H-score) over three runs are reported.} 
\label{fig1}
\end{figure}

\begin{figure*}
\centering
\includegraphics[width=16cm, height=6.8cm]{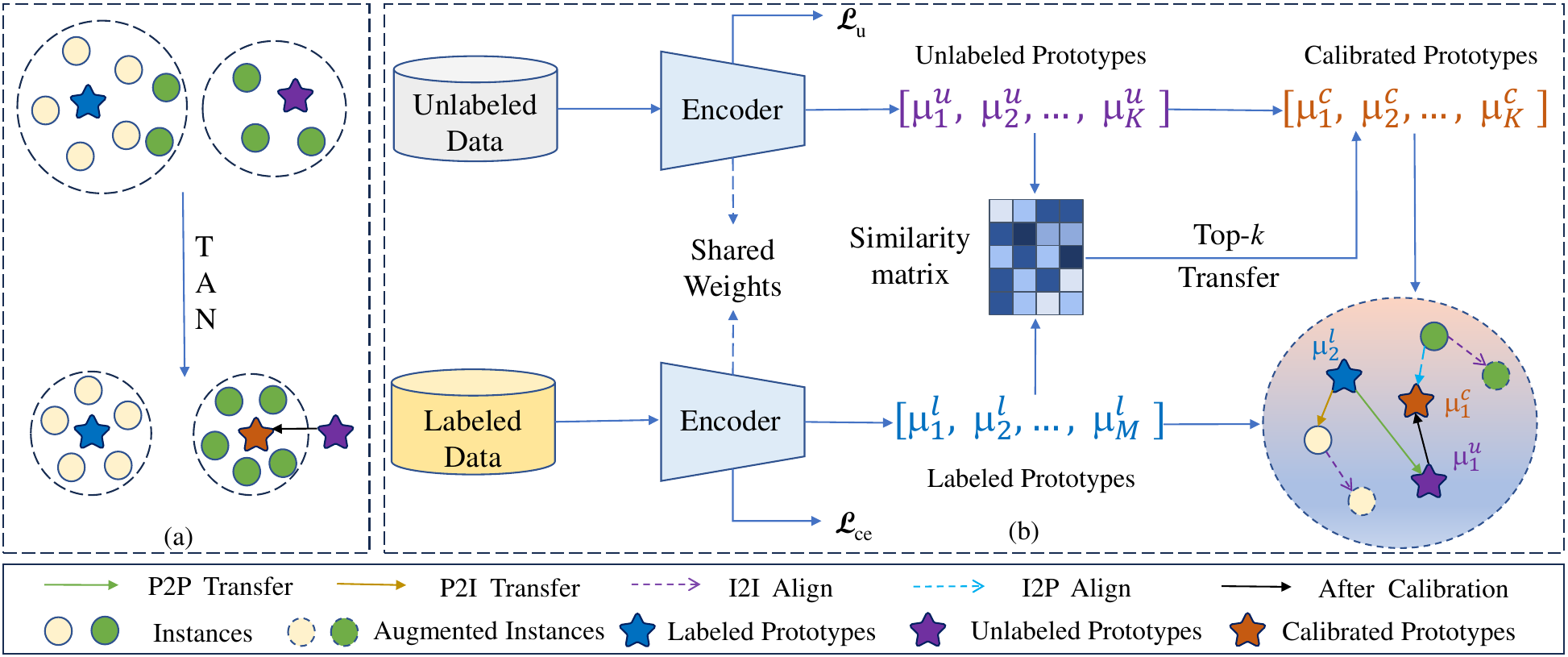}
\caption{(a) An illustration of the biased knowledge transfer (Top) and the effectiveness of our knowledge transfer mechanisms (Bottom). (b) An overview of our model.} 
\label{fig2}
\end{figure*}

Most existing works \citep{pretrain,dpn,diffusion} adopt a two-stage approach to address GCD: pre-training on labeled data and then transferring the pre-trained model for pseudo-label training on unlabeled data. Even though these methods have achieved good performance on known categories, they usually perform poorly on novel categories due to the lack of supervision (yellow bar in Fig. \ref{fig1}), which limits their applications in the real world. 
We attribute the poor performance to two reasons: \textbf{Biased knowledge transfer} and \textbf{Noisy representation learning}. 
First, models pretrained on labeled data with only known categories tend to be over-confident and biased towards known categories, so instances with novel categories in unlabeled data can be easily misclassified into known categories (Fig. \ref{fig2} (a) Top), where the biased knowledge transfer can lead to noisy estimation of learning targets for novel categories (e.g., biased category prototypes \citep{dpn} or unreliable pseudo labels \citep{thu2021}). 
Second, under the noisy learning targets, it is usually hard for current models to learn discriminative representations from unlabeled data effectively. Moreover, these methods solely focus on instance- \citep{gcd} or prototype-based discrimination \citep{dpn} with the noisy learning targets, but fail to combine them together to capture both instance- and category-level semantics, which can further disrupt the performance.

To mitigate above issues, we propose a \textit{\textbf{T}ransfer and \textbf{A}lignment \textbf{N}etwork} (TAN) to calibrate the biased knowledge and learn discriminative features. 
We first propose two knowledge transfer mechanisms to calibrate the biased knowledge caused by pre-training.
Specifically, we model different categories with prototypes \citep{proto}, then we leverage prototypes of known categories in labeled data as a prior to guide the training process on unlabeled data. 
On the one hand, we propose \textit{Prototype-to-Instance Transfer} (\textbf{P2I Trans}) to cluster instances with known categories in unlabeled data around these prototypes to form compact clusters, which can make different categories discriminative and avoid boundary overlap between known and novel categories. 
On the other hand, inspired by the fact that similar categories may share some common features (e.g., cats and dogs), we propose \textit{Prototype-to-Prototype Transfer} (\textbf{P2P Trans}) to transfer these prototypes to calibrate the noisy prototypes estimated from unlabeled data. To avoid negative transfer between dissimilar categories, we introduce semantic similarities between categories as weights and select only the most similar \textit{k} prototypes for the prototype calibration. By transferring knowledge from known to novel categories, the calibrated prototypes can be used as reliable learning targets to guide the subsequent representation learning.
Combining P2I and P2P Trans, TAN can learn clear decision boundaries for known categories and reliable prototypes for novel categories, which can help to alleviate the effects of biased knowledge transfer (Fig. \ref{fig2} (a) Bottom).

After knowledge transfer, we further propose two feature alignment mechanisms to acquire both instance- and category-level knowledge from unlabeled data with less noise.
First, we propose \textit{Instance-to-Prototype Alignment} (\textbf{I2P Align}) to pull instance features closer to the corresponding calibrated prototypes to acquire category-level knowledge (i.e., common category semantics embedded in prototypes from multiple instances), so that these features can be discriminative and compact around the prototypes to form more distinguishable decision boundaries. 
Second, we propose \textit{Instance-to-Instance Alignment} (\textbf{I2I Align}) to align instance features with their augmented features to acquire instance-level knowledge (i.e., specific instance semantics embedded in instance features), so that these features can be self-consistent and locally smooth for better representation learning \citep{simclr}.
As shown in Figure \ref{fig1}, our model achieves the best performance on three benchmark datasets, and the improved performance on novel categories further validates the effectiveness of our model. 
Last but not least, we theoretically justify the effectiveness of our model.

Our main contributions can be summarized as follows:
\begin{itemize}
  \item We propose a \textit{\textbf{T}ransfer and \textbf{A}lignment \textbf{N}etwork} (TAN) to mitigate the performance gap between known and novel categories in GCD.
  \item We propose two knowledge transfer mechanisms to alleviate the effects of biased knowledge transfer and two feature alignment mechanisms to acquire both instance- and category-level knowledge with less noise.
  \item Extensive experiments show that our model outperforms SOTA methods, and theoretical analysis further validates the effectiveness of our model.
\end{itemize}

\section{Related Work}
\subsection{Generalized Category Discovery}
Generalized Category Discovery (GCD) is a practical and challenging task formalised by \citet{gcd}. Under the open-world setting, GCD assumes that the newly collected unlabeled data contain novel categories that have never been seen during training. Due to the lack of supervision for novel categories, previous methods mainly employed pseudo-labeling based methods \citep{dtc,thu2020,mutual} or self-supervised methods \citep{ncl,simple,mtp} to learn from unlabeled data. For example, \citet{deepcluster} proposed to generate pseudo labels by clustering, \citet{thu2021} and \citet{ptjn} improved this method by generating more robust and consistent pseudo labels. As for self-supervised methods, \citet{gcd} proposed a semi-supervised k-means framework with contrastive learning to learn discriminative features, and \citet{dpn} proposed a decoupled prototypical network to decouple known and novel categories from unlabeled data.
Another line of work focuses on the discovery of novel fine-grained categories from coarsely-labeled data \citep{fcdc,dna,ner}, which can transfer knowledge between different label hierarchies.
Despite the improved overall performance, most of these methods performed poorly on novel categories because of the biased knowledge transfer and noisy representation learning, which may limit their applications in the real world.

\subsection{Transfer Learning}
Transfer learning aims at transferring knowledge from the source domain to boost model performance on the target domain \cite{transfer2,tail}. GCD is related to transfer learning since we need to transfer knowledge from known to novel categories. Most of current methods are based on the pre-training and fine-tuning paradigm to transfer knowledge implicitly by initializing model parameters \citep{thu2021}. For example, \citet{gcd} proposed to use contrastive learning to pre-train their model, and \citet{mtp} combined supervised learning and masked language modeling to initialize their model. 
However, we think this paradigm is sub-optimal for GCD because models pre-trained on labeled data tend to be biased towards known categories.

\section{Method}
In this section, we first formulate the Generalized Category Discovery (GCD) task. Then we introduce our \textit{\textbf{T}ransfer and \textbf{A}lignment \textbf{N}etwork} (TAN) in detail. Specifically, we first pre-train a feature encoder and learn category prototypes from both labeled and unlabeled data. Then we propose two knowledge transfer mechanisms to learn clear decision boundaries for known categories and reliable prototypes for novel categories, which can help to mitigate the effects of biased knowledge transfer. After knowledge transfer, we propose two feature alignment mechanisms to capture both instance- and category-level semantics to learn discriminative features. The framework of our model is shown in Fig. \ref{fig2} (b).

\subsection{Problem Formulation}
Models trained on a labeled dataset $\mathcal{D}^{l} = \{(x_{i},y_{i})|y_{i} \in \mathcal{Y}_{k}\}$ can recognize pre-defined known categories $\mathcal{Y}_{k}$ well. However, in the open world, the trained models may encounter unlabeled data $\mathcal{D}^{u} = \{x_{i}|y_{i} \in \{\mathcal{Y}_{k}, \mathcal{Y}_{n} \}\}$ that contain both known categories $\mathcal{Y}_{k}$ and novel categories $\mathcal{Y}_{n}$, which can make the models fail. To cope with this limitation, Generalized Category Discovery (GCD) requires models to recognize both known and novel categories based on $\mathcal{D}^{l}$ and $\mathcal{D}^{u}$, without any annotation for novel categories. We denote $M = |\mathcal{Y}_{k}|$ as the number of known categories and $K = |\mathcal{Y}_{k}| + |\mathcal{Y}_{n}|$ as the number of all categories.
Finally, model performance will be measured on a testing set $\mathcal{D}^{t} = \{(x_{i},y_{i})|y_{i} \in \{\mathcal{Y}_{k}, \mathcal{Y}_{n} \}\}$.

\subsection{Pre-training and Prototype Learning}
We use the pre-trained BERT \citep{bert} as our feature encoder $F_{\theta}$ to extract feature $z_{i} = F_{\theta}(x_{i})$ for the input $x_{i}$. To adapt the pre-trained model to the downstream GCD task, we use cross-entropy loss on labeled data and masked language modeling (mlm) loss \citep{pretrain} on unlabeled data to pre-train $F_{\theta}$, where the mlm loss can help to learn general knowledge and reduce model bias towards known categories.

To transfer knowledge between different categories and acquire reliable learning targets for unlabeled data, we model categories with prototypes \citep{proto} and learn two sets of prototypes from the labeled and unlabeled dataset, respectively.
For the labeled dataset, we take average of all instance features belonging to the same category as labeled prototypes $P^{l} = \{\mu_{j}^{l}\}_{j=1}^{M}$, where $\mu_{j}^{l} = \frac{1}{|\mathcal{D}^{l}_{j}|} \sum_{x_{i} \in \mathcal{D}^{l}_{j}} F_{\theta}(x_{i})$ and $\mathcal{D}^{l}_{j}$ is a set of labeled instances from the category $j$. For unlabeled data, we follow \citet{dpn} to perform clustering and utilize cluster centers as estimated unlabeled prototypes $P^{u} = \{\mu_{j}^{u}\}_{j=1}^{K}$, where $\mu_{j}^{u} = \frac{1}{|\mathcal{D}^{u}_{j}|} \sum_{x_{i} \in \mathcal{D}^{u}_{j}} F_{\theta}(x_{i})$ and $\mathcal{D}^{u}_{j}$ is a set of unlabeled instances belonging to the cluster $j$.

\subsection{Knowledge Transfer}
To transfer knowledge between known and novel categories and alleviate the effects of biased knowledge transfer, we propose two knowledge transfer mechanisms called \textit{Prototype-to-Prototype Transfer} (\textbf{P2P Trans}) and \textit{Prototype-to-Instance Transfer} (\textbf{P2I Trans}).
\subsubsection{P2P Trans.} Due to the lack of supervision and the model bias towards known categories, the estimated unlabeled prototypes can be noisy and biased, especially for novel categories. So directly performing prototypical learning \citep{proto} on these noisy prototypes can lead to inferior results. To mitigate this issue, we propose to transfer knowledge between categories and treat the labeled prototypes as unbiased estimation for known categories to calibrate the noisy unlabeled prototypes. To avoid negative transfer between dissimilar categories, we introduce semantic similarities between categories as weights and select only the most similar \textit{k} prototypes for the prototype calibration. Specifically, we measure the semantic similarity of two categories based on the Euclidean distance of the corresponding prototypes:
\begin{equation}
    S_{i} = \{- \Vert{\mu_{i}^{u} - \mu_{j}^{l}}\Vert_{2} \mid \mu_{j}^{l} \in P^{l} \}
\end{equation}
where $\mu_{i}^{u}$ is the $i$-th unlabeled prototype and $\mu_{j}^{l}$ is the $j$-th labeled prototype. Then we select the top-\textit{k} similar labeled prototypes as the transfer set $S_{i}^{\prime}$ for each unlabeled prototype:
\begin{equation}
    T_{i} = \{ j \mid S_{ij} \in top_{k} (S_{i}) \}
\end{equation}

\begin{equation}
    S_{i}^{\prime} = \{ S_{ij} \mid j \in T_{i}  \}
\end{equation}
where $S_{ij}$ is the $j$-th element of $S_{i}$. Then we calculate the transfer weights by normalizing similarities in the transfer set:
\begin{equation}
    w_{i} = softmax(S_{i}^{\prime}/\sqrt{dim(z)})
\end{equation}
where $dim(z)$ is the feature dimension. Then we use the transfer set and weights to calibrate each unlabeled prototype. The calibrated prototypes $P^{c} = \{\mu_{i}^{c}\}_{i=1}^{K}$ can be estimated as follows:
\begin{equation}
    \mu_{i}^{c} = \alpha \cdot \mu_{i}^{u} + (1-\alpha) \cdot \sum_{j \in T_{i}} w_{ij} \cdot \mu_{j}^{l}
\end{equation}
where $\alpha$ is a weighting factor, $w_{ij}$ is the $j$-th element of $w_{i}$. By using category similarities as weights, we can transfer knowledge from labeled prototypes to calibrate the noisy unlabeled prototypes, where the calibrated prototypes can be used as reliable learning targets for unlabeled data later.

\subsubsection{P2I Trans.} Since the labeled prototypes are learned from ground-truth labels, they can be viewed as unbiased estimation for known categories. So we further utilize these labeled prototypes to guide the training process of unlabeled data with known categories, based on the pseudo labels from clustering. Specifically, we need to first match the prototypes for known categories in labeled and unlabeled data. Following the assumption that the closest prototypes in the feature space represent the same category, we can find the optimal match function $\mathcal{P}$ through a bipartite matching algorithm introduced by \citet{dpn}, where $\mu_{i}^{l}$ and $\mu_{\mathcal{P}(i)}^{u}$ represent the same known category. Then we cluster unlabeled instances with known categories around the corresponding prototypes:
\begin{equation}
    \mathcal{L}_{p2i} = \frac{1}{N}\sum_{i=1}^{M}\sum_{x_{j} \in \mathcal{D}^{u}_{\mathcal{P}(i)}} \Vert{F_{\theta}(x_{j}) - \mu_{i}^{l}}\Vert_{2}
\end{equation}
where $N$ is the number of unlabeled data belonging to known categories based on clustering results. $\mathcal{D}^{u}_{\mathcal{P}(i)}$ is a set of unlabeled data belonging to the cluster $\mathcal{P}(i)$ (i.e., the $i$-th known category). In this way, we can form compact clusters and distinguishable decision boundaries for known categories.

\subsubsection{Discussion.} The knowledge transfer mechanisms are effective towards the problem of biased knowledge transfer in two aspects. First, we can calibrate the biased unlabeled prototypes and estimate more reliable prototypes through the P2P Trans (Sec. Prototype Calibration), which can help to correct instances that are biased to known categories. Second, we can form compact clusters and clear decision boundaries for known categories through the P2I Trans, which can also help to correct the biased instances. 
Combining P2P and P2I Trans, our model can learn more reliable prototypes and more distinguishable decision boundaries for different categories, which can make them more discriminative and alleviate the effects of biased knowledge transfer (Fig. 2 (a) Bottom).



\subsection{Feature Alignment}
After knowledge transfer, we further propose two feature alignment mechanisms called \textit{Instance-to-Prototype Alignment} (\textbf{I2P Align}) and \textit{Instance-to-Instance Alignment} (\textbf{I2I Align}) to learn discriminative features from unlabeled data.

\subsubsection{I2P Align. }
After P2P Trans, the calibrated prototypes are less noisy and can be treated as reliable learning targets for unlabeled data. So we propose to acquire category-level knowledge by pulling instance features of unlabeled data closer to the corresponding calibrated prototypes based on the clustering results, so that these features can be discriminative and compact around the prototypes to form more distinguishable decision boundaries for different categories:
\begin{equation}
    \mathcal{L}_{i2p} = -\frac{1}{|\mathcal{D}^{u}|}\sum_{i=1}^{K}\sum_{x_{j} \in \mathcal{D}^{u}_{i}} \Vert{F_{\theta}(x_{j}) - \mu_{i}^{c}}\Vert_{2}
\end{equation}
where $|\mathcal{D}^{u}|$ is the number of unlabeled instances, $\mathcal{D}^{u}_{i}$ is a set of unlabeled instances belonging to the cluster $i$.

\begin{table*}
\centering

\begin{tabular}{lccccccccc}
\toprule
\multirow{2}*{Method} & \multicolumn{3}{c}{BANKING} &\multicolumn{3}{c}{StackOverflow} & \multicolumn{3}{c}{CLINC}\\ 
\cmidrule(r){2-4}  \cmidrule(r){5-7}  \cmidrule(r){8-10}
            &H-score    &Known    &Novel    &H-score    &Known    &Novel    &H-score    &Known    &Novel \\
\midrule
DeepCluster &13.97  &13.94  &13.99  &19.10  &18.22  &14.80  &26.48  &27.34  &25.67  \\
DCN         &16.33  &18.94  &14.35  &29.22  &28.94  &29.51  &29.20  &30.00  &28.45  \\
DEC         &17.82  &20.36  &15.84  &25.99  &26.20  &25.78  &19.78  &20.18  &19.40  \\
KM-BERT     &21.08  &21.48  &20.70  &16.93  &16.67  &17.20  &34.05  &34.98  &33.16  \\
KM-GloVe    &29.25  &29.11  &29.39  &28.32  &28.60  &28.05  &51.62  &51.74  &51.50  \\
AG-GloVe    &30.47  &29.69  &31.29  &29.95  &28.49  &31.56  &44.16  &45.17  &43.20  \\
SAE         &37.77  &38.29  &37.27  &62.65  &57.36  &69.02  &45.74  &47.35  &44.24  \\

\midrule
Simple   &40.52  &49.96  &34.08  &57.53  &57.87  &57.20  &62.76  &70.60  &56.49  \\
Semi-DC  &47.40  &53.37  &42.63  &64.90  &63.57  &61.20  &73.41  &75.60  &71.34  \\
Self-Labeling    &48.19  &61.64  &39.56  &59.99  &78.53  &48.53  &61.29  &80.06  &49.65  \\
CDAC+    &50.28  &55.42  &46.01  &75.78  &77.51  &74.13  &69.42  &70.08  &68.77  \\
DTC      &52.13  &59.98  &46.10  &63.22  &80.93  &51.87  &68.71  &82.34  &58.95  \\
Semi-KM  &54.83  &73.62  &43.68  &61.43  &81.02  &49.47  &70.98  &89.03  &59.01  \\
DAC      &54.98  &69.60  &45.44  &63.64  &76.13  &54.67  &78.77  &89.10  &70.59  \\
GCD      &55.78  &75.16  &44.34  &64.63  &82.00  &53.33  &63.08  &89.64  &48.66  \\
PTJN     &60.69  &77.20  &50.00  &77.48  &72.80  &82.80  &83.34  &91.79  &76.32  \\
DPN      &60.73  &80.93  &48.60  &83.13  &85.29  &81.07  &84.56  &92.97  &77.54  \\
\midrule
TAN (OC) &65.49  &81.21  &54.87  &85.20  &84.80  &85.60  &86.20  &92.74 &80.53  \\
\textbf{TAN} (\textbf{Ours})        &\textbf{66.70}  &\textbf{81.97}  &\textbf{56.23}  &\textbf{86.64}  &\textbf{86.36}  &\textbf{86.93}  &\textbf{87.02}  &\textbf{93.39}  &\textbf{81.46}  \\
Improvement    &+5.97    &+1.04  &+6.23  &+3.51  &+1.07  &+4.13  &+2.46  &+0.42  &+3.92  \\

\bottomrule

\end{tabular}
\caption{Average results (\%) over 3 runs on the testing set. OC means over clustering.}
\label{table1}
\end{table*}

\subsubsection{I2I Align.}
In addition to I2P Align, we also propose Instance-to-Instance Alignment (I2I Align) to acquire instance-level knowledge from unlabeled data. Specifically, we use data augmentation methods to generate augmented instance $x_{i}^{\prime}$ for each instance $x_{i}$. Then we perform instance-wise contrastive learning \citep{infonce} to pull $x_{i}$ closer to $x_{i}^{\prime}$ and push other instances away from $x_{i}$, which can help to learn self-consistent and locally-smooth features for better representation learning \citep{pcl}:
\begin{equation}
\mathcal{L}_{i2i} = -\frac{1}{|\mathcal{D}^{u}|}\sum_{i=1}^{|\mathcal{D}^{u}|}log\frac{exp(z_{i} \cdot z_{i}^{\prime}/\tau)}{\sum_{j=1}^{2B}exp(z_{i} \cdot z_{j}^{\prime}/\tau))}
\end{equation}
where $z_{i}^{\prime}$ is the feature of $x_{i}^{\prime}$, $\tau$ is a temperature hyper-parameter, and $B$ is the batch size. We further add cross-entropy loss $\mathcal{L}_{u}$ on unlabeled data to learn more discriminative features with cluster ids as pseudo labels \citep{deepcluster}. 
To avoid catastrophic forgetting for knowledge acquired from labeled data and mitigate the effects of label noise in unlabeled data \citep{ptjn}, we also add cross-entropy loss $\mathcal{L}_{ce}$ on labeled data with ground-truth labels.

\subsubsection{Overall Loss.} The objective of our model is defined as:
\begin{equation}
\mathcal{L}_{TAN} = \mathcal{L}_{p2i} + \mathcal{L}_{i2p} + \mathcal{L}_{i2i} + \mathcal{L}_{u} + \beta \mathcal{L}_{ce}
\end{equation}
where $\beta$ is a weighting factor. By combining instance- and prototype-based learning, our model can capture both instance- and category-level semantics from unlabeled data, which can help to learn discriminative representations. 
In summary, our model can mitigate the problem of biased knowledge transfer by transferring knowledge from labeled to unlabeled data and alleviate the noisy representation learning problem by acquiring both instance- and category-level knowledge, which can help to learn clear decision boundaries for different categories and boost our model performance.

\subsection{Theoretical Analysis}
We formalize the error bound of our model with the theory of unsupervised domain adaptation \citep{proto3,theory}. Training on both labeled and unlabeled data, the classification error of GCD can be written as the linear weighted sum of errors on labeled and unlabeled data:

\begin{equation}
\epsilon(h)=\gamma \cdot \epsilon_{u}(h,\hat y^u) + (1-\gamma) \cdot \epsilon_{l}(h,y^l)
\label{ErrorTPN}
\end{equation}
where $h$ is a hypothesis, $\gamma$ is a weighting factor, $\epsilon_{u}(h,\hat y^u)=E_{x\sim \mathcal{D}^u}|h(x)-\hat y^u|$ and $\epsilon_{l}(h,y^l)=E_{x\sim \mathcal{D}^l}|h(x)-y^l|$ represent the error over the sample distribution of unlabeled data $\mathcal{D}^u$ with pseudo labels $\hat y^u$ and labeled data $\mathcal{D}^l$ with ground-truth labels $y^l$, respectively.
Then we want to analyse how close the error $\epsilon(h)$ is to an oracle error $\epsilon_{u}(h,y^{u})$ that evaluates the model learned on the unlabeled data with ground truth labels $y^{u}$. Following the analysis in \citet{proto3}, difference between the two losses can be bounded by the following Lemma.

\begin{lemma} 
Let $h$ be a hypothesis in class $\mathcal{H}$. Then
\begin{equation}
\left|\epsilon(h)-\epsilon_{u}(h,y^u)\right| \le(1-\gamma) (\frac{1}{2}d_{\mathcal{H}\Delta{H}}(\mathcal{D}^l,\mathcal{D}^u)+\lambda)+\gamma\rho
\end{equation}
where $d_{\mathcal{H}\Delta{H}}(\mathcal{D}^l,\mathcal{D}^u)= 2 \sup_{h,h' \in\mathcal{H}} |\epsilon_u(h,h')-\epsilon_l(h,h')|$ measures the domain discrepancy between the labeled and unlabeled data in the hypothesis space $\mathcal{H}$. $\lambda=\epsilon_l(h^*,y^l)+\epsilon_{u}(h^*,y^{u})$ is the total error on the labeled and unlabeled data with the joint optimal hypothesis $h^*$. And $\rho$ denotes the ratio of false pseudo labels for unlabeled data.
\end{lemma}

From the Lemma 1 we can see that the bound are decided by three terms. First, the term $\lambda$ is negligibly small with the joint optimal hypothesis $h^*$ and ground-truth labels, which can be disregard. Second, for the discrepancy term $d_{\mathcal{H}\Delta{H}}(\mathcal{D}^l,\mathcal{D}^u)$ that can be quantified by category-level discrepancy of prototypes \citep{proto3}, the main discrepancy of the labeled and unlabeled data is from the differences between known and novel categories.
Our \textbf{knowledge transfer mechanisms} can mitigate this discrepancy by transferring knowledge from known to novel categories, where the P2P Trans can help to estimate more reliable prototypes for novel categories and mitigate the discrepancy between known and novel categories. Third, the ratio of false pseudo labels $\rho$ can be gradually reduced by learning more discriminative representations during training. Our \textbf{feature alignment mechanisms} can capture both instance- and category-level semantics to learn more discriminative features, which can reduce the noise of pseudo labels (Sec. Accuracy of Pseudo Labels). In summary, our knowledge transfer and feature alignment mechanisms can help to tighten the bound in Lemma 1, which can prove the effectiveness of our model theoretically.

\begin{table}
\setlength\tabcolsep{10pt}
\centering
\begin{tabular}{lccc}
\toprule
Variant & H-score & Known & Novel \\
\midrule  
TAN & 66.70 & 81.97 & 56.23\\
\midrule
w/o \textit{P2I Trans}    & 65.46 & 80.46 & 55.18 \\
w/o \textit{P2P Trans}    & 63.45 & 82.54 & 51.53 \\
w/o $\mathcal{L}_{ce}$    & 62.70 & 81.45 & 50.96 \\
w/o \textit{I2I Align}    & 61.26 & 80.19 & 49.56 \\
w/o $\mathcal{L}_{u}$     & 60.11 & 82.50 & 47.28 \\
w/o \textit{I2P Align}    & 59.10 & 84.51 & 45.44 \\
\bottomrule

\end{tabular}
\caption{Ablation study with different model variants.}
\label{table2}
\end{table}

\section{Experiments}
\subsection{Experimental Setup}
\subsubsection{Datasets.}
We validate the effectiveness of our model on three benchmark datasets.
\textbf{BANKING} is an intent detection dataset in the bank domain \citep{banking}.
\textbf{StackOverflow} is a question classification dataset processed by \citet{stack}.  
\textbf{CLINC} is a text classification dataset from diverse domains \citep{clinc}. For each dataset, we randomly select 25\% categories as novel categories and 10\% data as labeled data.

\subsubsection{Comparison with SOTA.}
We compare the proposed model with various baselines and SOTA methods.

\noindent \textbf{Unsupervised Models.}\quad  (1) DeepCluster: Deep Clustering \citep{deepcluster}. (2) DCN: Deep Clustering Network \citep{dcn}. (3) DEC: Deep Embedding Clustering \citep{dec}. (4) KM-BERT: KMeans with BERT embeddings \citep{bert}. (5) KM-GloVe: KMeans \citep{km} with GloVe embeddings \citep{glove}. (6) AG-GloVe: Agglomerative Clustering \citep{ag} with GloVe embeddings. (7) SAE: Stacked Auto Encoder.

\noindent \textbf{Semi-supervised Models.}\quad (1) Simple: A Simple Parametric model \citep{simple}. (2) Semi-DC: Deep Clustering \citep{deepcluster} pretrained on labeled data. (3) Self-Labeling: Self-Labeling Framework \citep{selflabel}. (4) CDAC+: Constrained Adaptive Clustering \citep{thu2020}. (5) DTC: Deep Transfer Clustering \citep{dtc}. (6) Semi-KM: KMeans with BERT pretrained on labeled data. (7) DAC: Deep Aligned Clustering \citep{thu2021}. (8) GCD: Label Assignment with Semi-supervised KMeans \citep{gcd}. (9) PTJN: Robust Pseudo-label Training \citep{ptjn}. (10) DPN: Decoupled Prototypical Network \citep{dpn}.

\subsubsection{Evaluation Metrics.}
We measure model performance with clustering accuracy on the testing set with Hungarian algorithm \citep{hungarian}.
(1) \textbf{H-score}: harmonic mean of the accuracy for known and novel categories, which can avoid evaluation bias towards known categories \citep{hscore}.
(2) \textbf{Known}: accuracy for instances with known categories.
(3) \textbf{Novel}: accuracy for instances with novel categories.

\subsubsection{Implementation Details.}
We use the pretrained bert-base-uncased model \citep{huggingface} and adopt its suggested hyper-parameters. We only fine-tune the last four Transformer layers with AdamW optimizer. 
Early stopping is used during pretraining with wait patience 20. 
For hyper-parameters, $k$ is set to 5, $\alpha$ is set to 0.8, $\beta$ is set to 100 and $\tau$ is set to 0.07. 
Training epochs for StackOverflow, BANKING and CLINC dataset are set to \{10, 20, 20\}. The learning rate for pretraining and training is set to $5e^{-5}$ and $1e^{-5}$, respectively.
For masked language modeling, the mask probability is set to 0.15 following previous works. And SimCSE \citep{simcse} is used to generate augmented instances.

\begin{table}
\setlength\tabcolsep{4pt}
\centering
\begin{tabular}{cc}
\toprule
\textbf{Query Category}    & \textbf{Selected Categories} \\
\midrule  
Play\_Music &Play\_Music, Next\_Song, Whisper \\
Timezone & Timezone, Travel\_Alert, Time \\
\midrule
Timer & Alarm, Reminder\_Update, Time \\
Book\_Flight & Flight\_Status, Book\_Hotel, Car\_Rental \\
\bottomrule

\end{tabular}
\caption{Examples of the top-3 similar categories.}
\label{table3}
\end{table}

\subsection{Results and Discussion}

\subsubsection{Main Results. }
We show the results in Table \ref{table1}. From the results we can get following observations. 
First, our model gets the best performance on all evaluation metrics and datasets, which can show the effectiveness of our model. 
Second, our model achieves the best results on H-score (average \textbf{3.98\%} improvement), which means that our model can better balance the performance on known and novel categories and alleviate the effects of model bias towards known categories. 
Third, our model achieves the best performance on accuracy for known categories (average \textbf{0.84\%} improvement). Thanks to the knowledge transfer and feature alignment mechanisms, our model can form compact clusters and discriminative decision boundaries for known categories, which means that our model can boost model performance on novel categories without sacrificing the model performance on known categories.
Last but not least, our model achieves the best performance on accuracy for novel categories (average \textbf{4.76\%} improvement). We attribute the significant improvement to following reasons. First, our knowledge transfer mechanisms (\textit{P2I Trans} and \textit{P2P Trans}) can help to alleviate the effects of biased knowledge transfer by calibrating the noisy prototypes and forming clear decision boundaries for both known and novel categories, where the calibrated prototypes can be used as reliable learning targets for the subsequent training. Second, our feature alignment mechanisms (\textit{I2P Align} and \textit{I2I Align}) can alleviate the effects of noisy representation learning by acquiring instance- and category-level knowledge simultaneously. And under the guidance of the calibrated prototypes, our model can learn discriminative features to form compact clusters with less noise.

\subsubsection{Ablation Study. }
We inspect the contribution of different components to our model on the BANKING dataset in Table \ref{table2}. First, removing different components from TAN degrades model performance on novel categories and H-score, which can show the effectiveness of different components towards mitigating the model bias to known categories and boosting model performance on novel categories. Removing components related to representation learning (\textit{I2I Align}, \textit{I2P Align} and $\mathcal{L}_{u}$) has the greatest impact on the performance, which means that learning discriminative features is crucial for novel categories. Second, removing \textit{P2I Trans} and \textit{I2I Align} degrades model performance on known categories since they are responsible for learning compact and discriminative clusters for known categories. Even though removing some components (\textit{P2P Trans}, \textit{I2P Align} and $\mathcal{L}_{u}$) can improve the performance on known categories, they can also greatly exacerbate the model bias and degrade the performance on novel categories. In summary, our model can balance model performance on known and novel categories by mitigating the model bias and learning discriminative features.

\begin{figure}
\centering
\includegraphics[width=7.5cm, height=3.8cm]{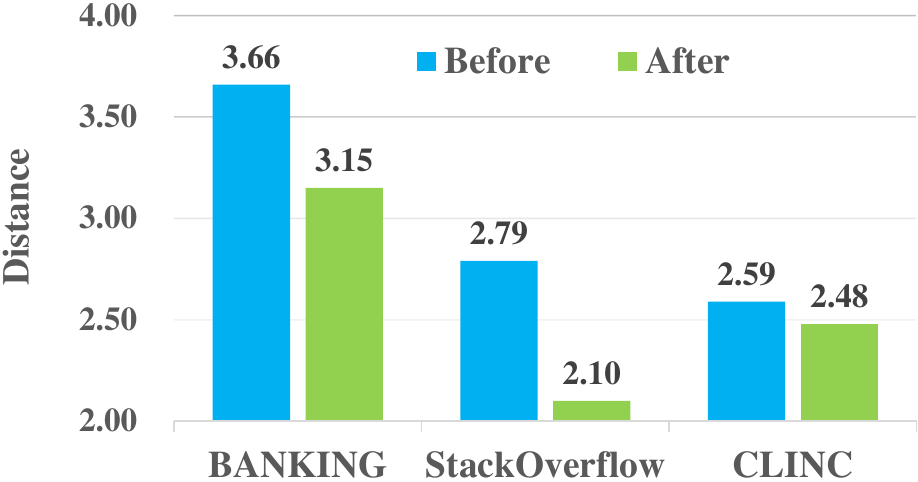}
\caption{Average distance between the ground-truth prototypes and the prototypes before and after calibration.} 
\label{fig3}
\end{figure}

\begin{table}
\centering
\begin{tabular}{lccc}
\toprule
Dataset & BANKING  & StackOverflow  & CLINC \\
\midrule
Ground Truth       & 77    & 20 & 150\\
\textbf{Ours}               & \textbf{74}    & \textbf{19} & \textbf{148}\\
\bottomrule
\end{tabular}
\caption{Estimation of the number of categories.}
\label{table4}
\end{table}

\subsubsection{Prototype Calibration. } \label{cali} We investigate the effectiveness of our Prototype Calibration (\textit{P2P Trans}) mechanism by answering the following questions.
\textbf{(1) Can prototype distances measure semantic similarities between categories?} In Table \ref{table3}, We show the top-3 selected known categories in Eq. (3) for both known (Top) and novel (Bottom) categories, based on the prototype distance. From the results we can see that the selected categories are highly relevant to query categories, which means that modeling categories with prototypes can preserve semantic similarities between categories. And by measuring distances between prototypes, we can transfer knowledge between similar categories and calibrate the noisy prototypes.
\textbf{(2) Can prototype calibration help to learn better prototypes?} In Fig. \ref{fig3}, we compare the average distance between the ground-truth prototypes and the prototypes before and after calibration. We can see that our model can learn prototypes that are closer to the ground-truth prototypes after calibration, which means that the prototype calibration can help to estimate more accurate and reliable prototypes.

\subsubsection{Real-world Applications. } In the real world, the number of categories $K$ is usually unknown. We show the robustness of our model towards this real-world setting from two aspects. \textbf{(1) Number of categories estimation.} We report the results on estimating the number of categories with the filtering algorithm \citep{thu2021} in Table \ref{table4}. We can see that our estimations come very close to the ground-truth number of categories, which can show the effectiveness of our model. \textbf{(2) Over Clustering. } To investigate the sensitivity of our model to the number of categories, we over-estimate the number of categories used for training and testing by a factor of one point two. As shown in Table \ref{table1}, our model (TAN (OC)) gets close performance even without knowing the ground-truth number of categories, which can show the robustness of our model towards the real-world settings.

\subsubsection{Accuracy of Pseudo Labels. } We report the accuracy of pseudo labels generated by different models for unlabeled data in Table \ref{table5}. Our model gets the highest accuracy, so the ratio of false pseudo labels $\rho$ in the Sec. Theoretical Analysis can be controlled by our model, which can verify the validity of our theoretical analysis.

\subsubsection{Visualization. } 
\begin{figure}
\centering
\subfigure[Before Training]{
\begin{minipage}[t]{0.5\linewidth}
\centering
\includegraphics[width=1.3in,height=2.5cm]{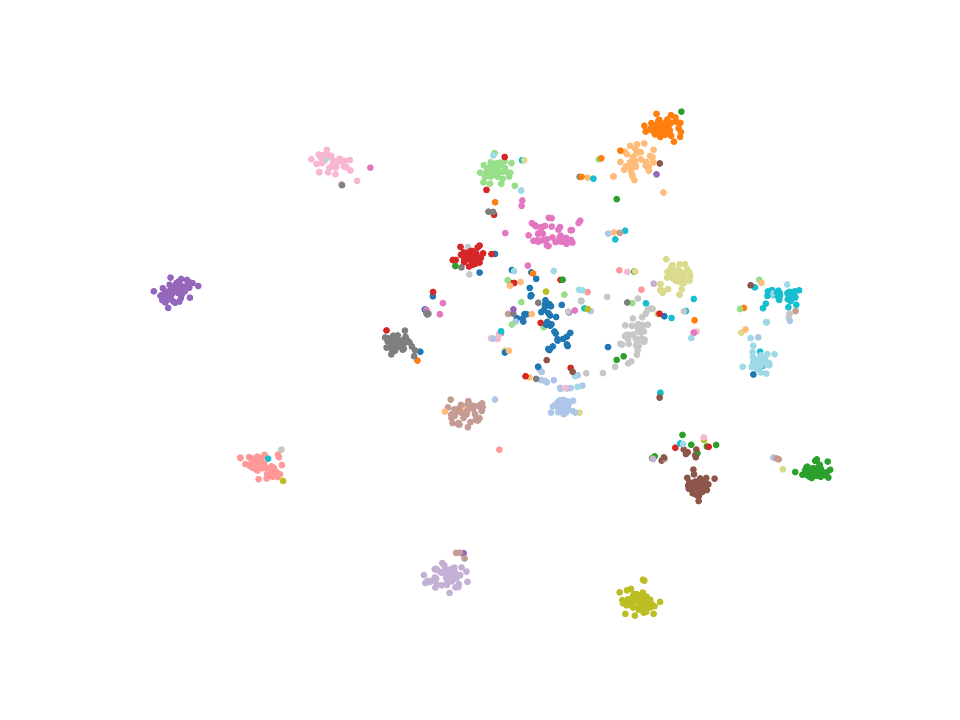}
\end{minipage}%
} \hspace{-10mm}
\subfigure[After Training]{
\begin{minipage}[t]{0.5\linewidth}
\centering
\includegraphics[width=1.3in,height=2.5cm]{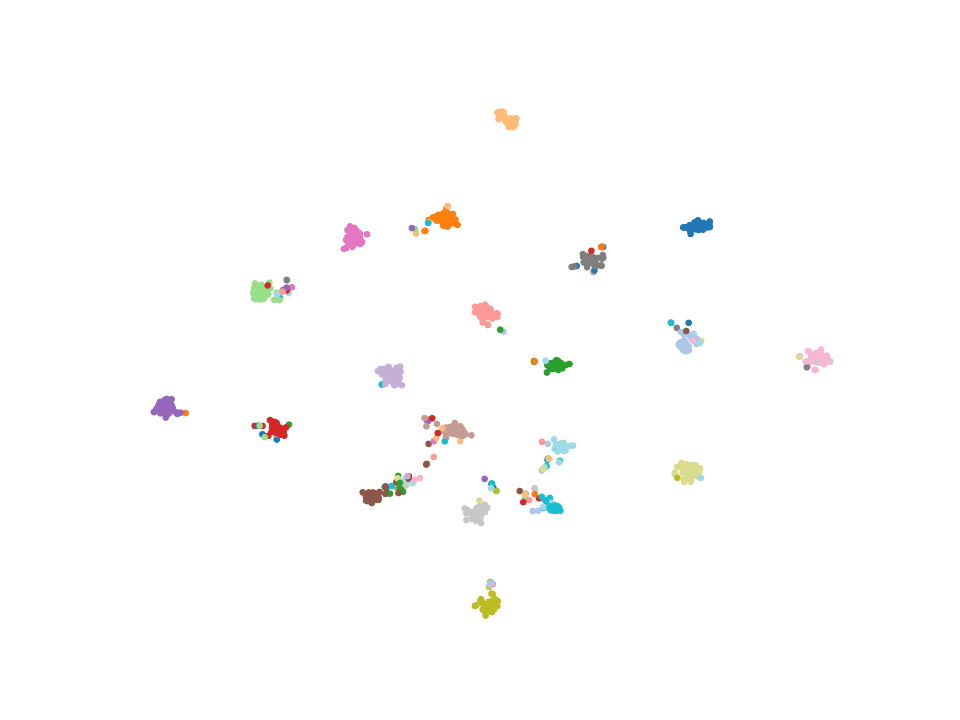}
\end{minipage}%
}%
\centering
\caption{The t-SNE visualization of embeddings.}
\label{fig4}
\end{figure}

We visualize the learned embeddings of our model before and after training with t-SNE in Fig. \ref{fig4}. From the figure we can see that novel categories are mixed together before training. And the clusters are more distinguishable after training, especially for novel categories, which indicates that our model can learn discriminative features and form distinguishable decision boundaries for different categories.

\begin{table}
\centering
\begin{tabular}{lccc}
\toprule
Method & CLINC  & StackOverflow & BANKING \\
\midrule
PTJN &80.23 &72.32  &67.50   \\
DPN  &80.89 &77.16  &68.34   \\
\textbf{TAN (Ours)} & \textbf{82.67} & \textbf{81.37} & \textbf{69.05}   \\ 
\bottomrule
\end{tabular}
\caption{Accuracy of pseudo labels.}
\label{table5}
\end{table}

\section{Conclusion}
In this paper, we propose \textit{\textbf{T}ransfer and \textbf{A}lignment \textbf{N}etwork} for GCD, which incorporates two knowledge transfer mechanisms to mitigate the effects of biased knowledge transfer and two feature alignment mechanisms to learn discriminative features with less noise.
By modeling different categories with prototypes and transferring knowledge from labeled to unlabeled data, our model can calibrate the noisy prototypes for novel categories and learn more discriminative clusters for known categories, which can help to mitigate the model bias towards known categories.
After knowledge transfer, our model can acquire both instance- and category-level knowledge by aligning instance features with both augmented features and the calibrated prototypes, which can help to learn more discriminative features and form more distinguishable decision boundaries for different categories.
Experimental results on three benchmark datasets show that our model outperforms SOTA methods, especially for model performance on novel categories. And the theoretical analysis further justifies the effectiveness of our model.

\section*{Acknowledgments}
This work was supported by National Key Research and Development Program of China (2022ZD0117102), National Natural Science Foundation of China (62293551, 62177038, 62277042, 62137002, 61721002, 61937001, 62377038). Innovation Research Team of Ministry of Education (IRT\_17R86), Project of China Knowledge Centre for Engineering Science and Technology, "LENOVO-XJTU" Intelligent Industry Joint Laboratory Project.

\bibliography{aaai24}
\end{document}